\documentclass[10pt,journal,compsoc]{IEEEtran}
\usepackage[utf8]{inputenc} 
\usepackage[T1]{fontenc}    
\usepackage{hyperref}       
\usepackage{url}            
\usepackage{booktabs}       
\usepackage{amsfonts}       
\usepackage{nicefrac}       
\usepackage{microtype}      
\usepackage{algorithm}
\usepackage{algpseudocode}
\usepackage{amsmath}
\usepackage{amsthm}
\usepackage{amssymb}
\usepackage{siunitx}
\usepackage{textcomp}
\usepackage{forloop}
\usepackage{mathtools}
\usepackage{graphicx}
\usepackage{comment}
\usepackage{xcolor}
\usepackage[english]{babel}
\graphicspath{{figures/}}
\DeclareGraphicsExtensions{.pdf,.png,.jpg,.mps,.eps,.ps}

\newtheorem{theorem}{Theorem}[section]
\newtheorem{corollary}{Corollary}[theorem]

\newcommand\observationSpace{\mathfrak{Y}}
\usepackage{xpatch}
\definecolor{c:recon}{rgb}{0.15,0.35,0.5}
\definecolor{c:ent}{rgb}{0.1,0.5,0.2}
\definecolor{c:dyn}{rgb}{0.2,0,0.4}

\newcommand{\defvec}[1]{\expandafter\newcommand\csname v#1\endcsname{{\mathbf{#1}}}}
\newcounter{ct}
\forLoop{1}{26}{ct}{
    \edef\letter{\alph{ct}}
    \expandafter\defvec\letter
}
\forLoop{1}{26}{ct}{
    \edef\letter{\Alph{ct}}
    \expandafter\defvec\letter
}

\DeclareMathOperator{\KL}{\mathbb{D}_{\text{KL}}}

\newcommand{\bm}[1]{\boldsymbol{\mathbf{#1}}}

\allowdisplaybreaks

\begin{document}
\title{Streaming Variational Monte Carlo}

%
%
%
%
\author{Yuan~Zhao*\thanks{* equal contribution},~Josue~Nassar*,~Ian~Jordan,~M\'{o}nica~Bugallo,~and~Il~Memming~Park

\IEEEcompsocitemizethanks{\IEEEcompsocthanksitem All authors are with Stony Brook University, Stony Brook, NY, 11794.
	Y.~Zhao and I.~M.~Park are with the Department of Neurobiology and Behavior.
	J.~Nassar and M.~Bugallo are with the Department of Electrical and Computer Engineering.
	I.~Jordan is with the Department of Applied Mathematics and Statistics.}
}

\IEEEtitleabstractindextext{%
	\begin{abstract}
		Nonlinear state-space models are powerful tools to describe dynamical structures in complex time series.
		In a streaming setting where data are processed one sample at a time, simultaneous inference of the state and its nonlinear dynamics has posed significant challenges in practice.
		We develop a novel online learning framework, leveraging variational inference and sequential Monte Carlo, which enables flexible and accurate Bayesian joint filtering.
		Our method provides an approximation of the filtering posterior which can be made arbitrarily close to the true filtering distribution for a wide class of dynamics models and observation models.
		Specifically, the proposed framework can efficiently approximate a posterior over the dynamics using sparse Gaussian processes, allowing for an interpretable model of the latent dynamics.
		Constant time complexity per sample makes our approach amenable to online learning scenarios and suitable for real-time applications.
	\end{abstract}

	}

\maketitle

\IEEEdisplaynontitleabstractindextext

%
\IEEEpeerreviewmaketitle

\IEEEraisesectionheading{\section{Introduction}\label{sec:introduction}}
Nonlinear state-space models are generative models for complex time series with underlying nonlinear dynamical structure~\cite{Haykin1998,Ko2009,Mattos2016}.
Specifically, they represent nonlinear dynamics in the latent state-space, $x_t$, that capture the spatiotemporal structure of noisy observations, $y_t$:
\begin{subequations}\label{eq:ssm}
	\begin{align}
		x_t & = f_{\theta}(x_{t-1}, u_t) + \epsilon_t
		\,  & \text{(state dynamics model)}
		\label{eq:sde}
		\\
		y_t & \sim P(y_t \vert g_\psi(x_t)) \,         & \text{(observation model)}
		\label{eq:obs}
	\end{align}
\end{subequations}
where $f_{\theta}$ and $g_{\psi}$ are continuous vector functions, $P$ denotes a probability distribution, and $\epsilon_t$ is intended to capture unobserved perturbations of the state $x_t$.
Such state-space models have many applications (e.g., object tracking) where the flow of the latent states is governed by known physical laws and constraints or where learning an interpretable model of the laws is of great interest, especially in neuroscience~\cite{Roweis2001,Sussillo2013,Frigola2014,Daniels2015,Zhao2016,Nassar2019}.
If the parametric form of the model and the parameters are known a priori, then the latent states $x_t$ can be inferred online through the filtering distribution, $p(x_t \vert \vy_{1:t})$, or offline through the smoothing distribution, $p(\vx_{1:t} \vert \vy_{1:t})$~\cite{Ho1964,Sarkka2013}. Otherwise the challenge is in learning the parameters of the state-space model, $\{\theta, \psi\}$, which is known in the literature as the system identification problem.

In a streaming setting where data is processed one sample at a time, joint inference of the state and its nonlinear dynamics has posed significant challenges in practice.
In this study, we are interested in online algorithms that can recursively solve the dual estimation problem of learning both the latent trajectory, $\vx_{1:t}$, in the state-space and the parameters of the model, $\{\theta, \psi\}$, from streaming observations~\cite{Haykin2001}.

Popular solutions such, as the extended Kalman filter (EKF) or the unscented Kalman filter (UKF)~\cite{wan2000unscented}, build an online dual estimator using nonlinear Kalman filtering by augmenting the state-space with its parameters~\cite{wan2000unscented,Wan2001,wan1999dual,wan1997dual}. While powerful, they usually provide coarse approximations to the filtering distribution and involve many hyperparameters to be tuned which hinder their practical performance.
Moreover, they do not take advantage of modern stochastic gradient optimization techniques commonly used throughout machine learning and are not easily applicable to arbitrary observation likelihoods.

Recursive stochastic variational inference has been proposed for streaming data assuming either independent~\cite{Broderick2013} or temporally-dependent samples~\cite{Frigola2014,Zhao2017,campbell2021online}.
However the proposed variational distributions are not guaranteed to be good approximations to the true posterior.
As opposed to variational inference, sequential Monte Carlo (SMC) leverages importance sampling to build an approximation to the target distribution in a data streaming setting~\cite{doucet2009tutorial,Doucet2013}.
However, its success heavily depends on the choice of proposal distribution and the (locally) optimal proposal distribution usually is only available in the simplest cases~\cite{doucet2009tutorial}.
While work has been done on learning good proposals for SMC~\cite{cornebise2014adaptive, gu2015neural,guarniero2017iterated,Naesseth2018} most are designed only for offline scenarios targeting the smoothing distributions instead of the filtering distributions.
In~\cite{cornebise2014adaptive}, the proposal is learned online but the class of dynamics for which this is applicable to is extremely limited.

In this paper, we propose a novel sequential Monte Carlo method for inferring a state-space model for the streaming time series scenario that adapts the proposal distribution on-the-fly by optimizing a surrogate lower bound to the log normalizer of the filtering distribution.
Moreover, we choose the sparse Gaussian process (GP)~\cite{Titsias2009} for modeling the unknown dynamics that allows for $\mathcal{O}(1)$ recursive Bayesian inference.
Specifically our contributions are:
\begin{enumerate}
	\item We prove that our approximation to the filtering distribution \textbf{converges to the true filtering distribution}.
	\item Our objective function allows for \textbf{unbiased gradients} which lead to improved performance.
	\item To the best of our knowledge, we are the first to use particles to represent the posterior of inducing variables of the sparse Gaussian processes, which allows for \textbf{accurate Bayesian inference} on the inducing variables rather than the typical variational approximation and closed-form weight updates.
	\item Unlike many efficient filtering methods that usually assume Gaussian or continuous observations, our method allows \textbf{arbitrary~observational~distributions}.
\end{enumerate}

\section{Streaming Variational Monte Carlo}

Given the state-space model defined in~\eqref{eq:ssm}, the goal is to obtain the latent state, $x_t \in \mathbb{R}^{d_x}$, given a new observation, $y_t \in \observationSpace$, where $\observationSpace$ is a measurable space (typically $\observationSpace = \mathbb{R}^{d_y}$ or $\observationSpace = \mathbb{N}^{d_y}$).
Under the Bayesian framework, this corresponds to computing the filtering posterior distribution at time $t$
\begin{equation}\label{eq:filtering_distribution}
	p(x_t \vert \vy_{1:t})
	=
	\frac{p(y_t \vert x_t)}{p(y_t \vert \vy_{1:t-1})} p(x_t \vert \vy_{1:t-1})
\end{equation}
which recursively uses the previous filtering posterior distribution, $p(x_t \vert \vy_{1:t-1}) = \int p(x_t\vert x_{t-1})p(x_{t-1}\vert \vy_{1:t-1})dx_{t-1}$.

However, the above posterior is generally intractable except for limited cases~\cite{Haykin2001} and thus we turn to approximate methods.
Two popular approaches for approximating~\eqref{eq:filtering_distribution} are sequential Monte Carlo~(SMC)~\cite{doucet2009tutorial}  and variational inference~(VI)~\cite{blei2017variational,zhang2018advances,wainwright2008graphical}.
In this work, we propose to combine sequential Monte Carlo and variational inference, which allows us to utilize modern stochastic optimization while leveraging the flexibility and theoretical guarantees of SMC. We refer to our approach as \textit{streaming variational Monte Carlo} (SVMC).
For clarity, we review SMC and VI in the follow sections.

\subsection{Sequential Monte Carlo}
SMC is a sampling based approach to approximate Bayesian inference that is designed to recursively approximate a sequence of distributions $p(\vx_{0:t}\vert \vy_{1:t})$ for $t=1, \ldots$, using samples from a proposal distribution, $r(\vx_{0:t} \vert \vy_{1:t}; \bm{\lambda}_{0:t})$ where $\bm{\lambda}_{0:t}$ are the parameters of the proposal~\cite{doucet2009tutorial}.
Due to the Markovian nature of the state-space model in~\eqref{eq:ssm}, the smoothing distribution, $p(\vx_{0:t} \vert \vy_{1:t})$, can be expressed as
\begin{equation}\label{eq:smoothing}
	p(\vx_{0:t} \vert \vy_{1:t}) \propto p(x_0) \prod_{j=1}^t p(x_t \vert x_{t-1}) p(y_t \vert x_t).
\end{equation}
We enforce the same factorization for the proposal, $r(\vx_{0:t} \vert \vy_{1:t}; \bm{\lambda_{0:t}}) = r_0(x_0; \lambda_0)\prod_{j=1}^t r_j(x_j \vert x_{j-1}, y_j; \lambda_j)$.

A naive approach to approximating~\eqref{eq:smoothing} is to use standard importance sampling (IS)~\cite{mcbook}.
$N$ samples are sampled from the proposal distribution, $\vx_{0:t}^1, \cdots, \vx_{0:t}^N \sim r(\vx_{0:t}; \bm{\lambda}_{0:t})$, and are given weights according to
\begin{equation}\label{eq:naive_is}
	w^i_{0:t} = \frac{p(x_0^i)\prod_{j=1}^t p(x_j^i\vert x_{j-1}^i) p(y_j \vert x_j^i) }{r_0(x_0^i; \lambda_0) \prod_{j=1}^t r_j(x_j^i \vert x_{j-1}^i, y_j; \lambda_j)}.
\end{equation}
The importance weights can also be computed recursively
\begin{equation}
	w_{0:t}^i =\prod_{s=0}^t w^i_s,
\end{equation}
where
\begin{equation}\label{eq:sequential_iw}
	w_s^i = \frac{p(y_s \vert x_s^i)p(x_s^i \vert x_{s-1}^i)}{r_s(x_s^i \vert x_{s-1}^i, y_s; \lambda_s)}.
\end{equation}
The samples and their corresponding weights, $\{ (\vx_{0:t}^i, w_{0:t}^i) \}_{i=1}^N$, are used to build an approximation to the target distribution
\begin{equation}
	p(\vx_{0:t} \vert \vy_{1:t}) \approx \hat{p}(\vx_{0:t} \vert \vy_{1:t}) = \sum_{i=1}^N \frac{w_{0:t}^i}{\sum_\ell w_{0:t}^\ell}\delta_{\vx_{0:t}^i}
\end{equation}
where $\delta_x$ is the Dirac-delta function centered at $x$.
While straightforward, naive IS suffers from the weight degeneracy issue; as the length of the time series, $T$, increases all but one of the importance weights will go to 0~~\cite{doucet2009tutorial}.

To alleviate this issue, SMC leverages sampling-importance-resampling (SIR).
Suppose at time $t-1$, we have the following approximation to the smoothing distribution
\begin{equation}
	\hat{p}(\vx_{0:t-1} \vert \vy_{1:t-1}) = \frac{1}{N} \sum_{i=1}^N \frac{w_{t-1}^i}{\sum_\ell w_{t-1}^\ell}\delta_{\vx_{0:t-1}^i},
\end{equation}
where $w_{t-1}^i$ is computed according to~\eqref{eq:sequential_iw}.
Given a new observation, $y_t$, SMC starts by resampling ancestor variables, $a_t^i \in \{1, \ldots, N \}$ with probability proportional to the importance weights, $w_{t-1}^j$.
$N$ samples are then drawn from the proposal, $x_t^i \sim r_t(x_t \vert x_{t-1}^{a_t^i}, y_t; \lambda_t)$, and their importance weights are computed, $w_t^i$, according to~\eqref{eq:sequential_iw}.
The introduction of resampling allows for a (greedy) solution to the weight degeneracy problem.
Particles with high weights are deemed good candidates and are propagated forward while the ones with low weights are discarded.

The updated approximation to $p(\vx_{0:t} \vert \vy_{1:t})$ is now
\begin{equation}\label{eq:smoothing_approx}
	\hat{p}(\vx_{0:t} \vert \vy_{1:t}) = \frac{1}{N} \sum_{i=1}^N \frac{w_{t}^i}{\sum_\ell w_{t}^\ell}\delta_{\vx_{0:t}^i},
\end{equation}
where $\vx_{0:t}^i = (x_t^i, \vx_{0:t-1}^{a_t^i})$.
Marginalizing out $\vx_{0:t-1}$ in \eqref{eq:smoothing_approx} gives an approximation to the filtering distribution:
\begin{equation}\label{eq:filtering_approx}
	\begin{split}
		p(x_t \vert \vy_{1:t})
		&= \int p(\vx_{0:t} \vert \vy_{1:t}) \mathrm{d}\vx_{0:t-1}
		\\
		&\approx \int \sum_{i=1}^N \frac{w_t^i}{\sum_\ell w_t^\ell} \delta_{\vx_{0:t}^i} \\
		& = \sum_{i=1}^N \frac{w_t^i}{\sum_\ell w_t^\ell} \delta_{x_{t}^i}.
	\end{split}
\end{equation}
As a byproduct, the weights produced in an SMC run yield an unbiased estimate of the marginal likelihood of the smoothing distribution~\cite{Doucet2013}
\begin{equation}\label{eq:marg_like_estimate}
	\mathbb{E}[\hat{p}(\vy_{1:t})] =
	\mathbb{E}
	\left[
		\prod_{s=1}^t \frac{1}{N} \sum_{i=1}^N w_s^i
		\right]
	= p(\vy_{1:t}),
\end{equation}
and a biased but \textit{consistent} estimate of the marginal likelihood of the filtering distribution~\cite{Doucet2013,van2000asymptotic}
\begin{equation}\label{eq:marg_filter_like_estimate}
	\mathbb{E}[\hat{p}(y_t \vert \vy_{1:t-1})]
	= \mathbb{E}\left[ \frac{1}{N}\sum_{i=1}^N w_t^i \right].
\end{equation}
For completeness, we reproduce the consistency proof of \eqref{eq:marg_filter_like_estimate} in section~\ref{proof:consistent} of the appendix.
The recursive nature of SMC makes it constant complexity per time step and constant memory because only the samples and weights generated at time $t$ are needed, $\{w_t^i, x_t^i\}_{i=1}^N$, making them a perfect candidate to be used in an online setting~\cite{adali2010adaptive}.
These attractive properties have allowed SMC to enjoy much success in fields such as robotics~\cite{thrun2002particle}, control engineering~\cite{greenfield2003adaptive} and target tracking~\cite{gustafsson2002particle}.

The success of an SMC sampler crucially depends on the design of the proposal distribution, $r_t(x_t \vert x_{t-1}, y_t;\lambda_t)$.
A common choice for the proposal distribution is the transition distribution, $r_t(x_t \vert x_{t-1}, y_t;\lambda_t) = p(x_t \vert x_{t-1})$, which is known as the bootstrap particle filter (BPF)~\cite{gordon1993novel}. While simple, it is well known that the BPF needs a large number of particles to perform well and suffers in high-dimensions~\cite{bickel2008sharp}.
In addition, BPF requires the knowledge of $p(x_t \vert x_{t-1})$ which may not be known.

Designing a proposal is even more difficult in an online setting because a proposal distribution that was optimized for the system at time $t$ may not be the best proposal $K$ steps ahead.
For example, if the dynamics were to change abruptly, a phenomenon known as concept drift~\cite{vzliobaite2016overview}, the previous proposal may fail for the current time step.
Thus, we propose to adapt the proposal distribution online using variational inference.
This allows us to utilize modern stochastic optimization to adapt the proposal on-the-fly while still leveraging the theoretical guarantees of SMC.

\subsection{Variational Inference}
In contrast to SMC, VI takes an optimization approach to approximate Bayesian inference.
In VI, we approximate the target posterior, $p(x_t \vert \vy_{1:t})$, by a class of simpler distributions, $q(x_t; \vartheta_t)$, where $\vartheta_t$ are the parameters of the distribution.
We then minimize a divergence (which is usually the Kullback-Leibler divergence (KLD)) between the posterior and the approximate distribution in the hopes of making $q(x_t ; \vartheta_t)$ closer to $p(x_t \vert \vy_{1:t})$.
If the divergence used is KLD, then minimizing the KLD between these distributions is equivalent to maximizing the so-called evidence lower bound (ELBO)~\cite{wainwright2008graphical,blei2017variational}:
\begin{align}\label{eq:elbo}
	\mathcal{L} & (\vartheta_t)
	= \mathbb{E}_q[\log p(x_t, \vy_{1:t}) - \log q(x_t; \vartheta_t) ],
	\\
	\nonumber
	            & = \mathbb{E}_q[
	\log \mathbb{E}_{p(x_{t-1} \vert \vy_{1:t-1})}
	[p(x_t, x_{t-1}, \vy_{1:t})]
	- \log q(x_t; \vartheta_t)
	].
\end{align}
For filtering inference, the intractability introduced by marginalizing over $p(x_{t-1} \vert \vy_{1:t-1})$ in \eqref{eq:elbo} makes the problem much harder to optimize, rendering variational inference impractical in a streaming setting where incoming data are temporally dependent.

\noindent
\begin{algorithm}
	\caption[]{Streaming Variational Monte Carlo (Step $t$)}
	\label{alg:smc}
	\begin{algorithmic}[1]
		\Require $\{x_{t-1}^i, w_{t-1}^i\}_{i=1}^N, \Theta_{t-1}, y_t, \alpha$
		\For{$k = 1, \ldots, N_{\text{SGD}}$}
		\For{$i = 1, \ldots, L$}
		\State $a_t^i \sim \Pr(a_{t}^i = j) \propto w_{t-1}^j$ \Comment {\textit{Resample}}
		\State $x_t^i \sim r(x_t \vert x_{t-1}^{a_t^i}, y_t; \Theta_{t-1})$ \Comment {\textit{Propose}}
		\State $w_t^i \gets \frac{p(x_t^i \vert x_{t-1}^{a_t^i}; \Theta_{t-1}) p(y_t \vert x_{t-1}^{a_t^i}; \Theta_{t-1})}{r(x_t^i \vert x_{t-1}^{a_t^i}, y_t; \Theta_{t-1})}$ \Comment {\textit{Weigh}}
		\EndFor
		\State $\tilde{\mathcal{L}_t} \gets \log(\sum_{i}  w_t^i)$
		\State $\Theta_t \gets \Theta_{t-1} + \alpha \nabla_{\Theta} \tilde{\mathcal{L}_t}$ \Comment{SGD}
		\EndFor
		\State Resample, propose and reweigh $N$ particles\\
		\Return $\{x_t^i, w_{t}^i\}_{i=1}^N$, $\Theta_t$
	\end{algorithmic}
\end{algorithm}

\subsection{A Tight Lower Bound}
Due to the intractability of the filtering distribution, the standard ELBO is difficult to optimize forcing us to define a different objective function.
As stated above, we know that the sum of importance weights is an unbiased estimator of $p(\vy_{1:t})$.
Jensen's inequality applied to \eqref{eq:marg_like_estimate}~\cite{Naesseth2018,anh2018autoencoding} gives,
\begin{equation}\label{eq:jensen}
	\log p(\vy_{1:t})
	=
	\log \mathbb{E}[\hat{p}(\vy_{1:t})]
	\geq
	\mathbb{E}[\log \hat{p}(\vy_{1:t})].
\end{equation}
Expanding \eqref{eq:jensen}, we obtain
\begin{gather}
	\begin{split}
		\log p(y_{t} &\vert \vy_{1:t-1}) + \log p(\vy_{1:t-1})
		\\
		& \qquad \geq \mathbb{E}[\log \hat{p}(y_{t} \vert \vy_{1:t-1})]
		+ \mathbb{E}[\log \hat{p}(\vy_{1:t-1})],
	\end{split}
	\\[2ex]
	\log p(y_{t} \vert \vy_{1:t-1})
	\geq \mathbb{E}[\log \hat{p}(y_{t} \vert \vy_{1:t-1})] - \mathcal{R}_t(N)
\end{gather}
where $\mathcal{R}_t(N) = \log p(\vy_{1:t-1}) - \mathbb{E}[\log \hat{p}(\vy_{1:t-1})] \geq 0$ is the variational gap.
Leveraging this we propose to optimize
\begin{equation}\label{eq:filtering_elbo}
	\begin{split}
		\widetilde{\mathcal{L}}_t(\Theta_t)
		= \mathbb{E}[\log \hat{p}(y_{t} \vert \vy_{1:t-1})]
		- \mathcal{R}_t(N),
		\\
		= \mathbb{E}\left[
			\log\big( \sum_{i=1}^N  w_t^i \big)
			\right] - \mathcal{R}_t(N).
	\end{split}
\end{equation}
We call $\widetilde{\mathcal{L}}_t$ the \textit{filtering ELBO}; it is a lower bound to the log normalization constant (log partition function) of the filtering distribution where $\mathcal{R}_t(N)$ accounts for the bias of the estimator~\eqref{eq:marg_filter_like_estimate}.
As $\mathcal{R}_t(N)$ is \textbf{not} a function of $\Theta_{t}$, it can be ignored when computing gradients.

There exists an implicit posterior distribution that arises from performing SMC given by~\cite{Cremer2017},
\begin{align}\label{eq:implicit_approx}
	\tilde{q}(x_t & \vert \vy_{1:t}) = p(x_{t}, \vy_{1:t})\mathbb{E}\left[\frac{1}{\hat{p}(\vy_{1:t})}\right],
	\\
	\nonumber
	              & = p(x_t, y_t \vert \vy_{1:t-1}) \mathbb{E}\left[\hat{p}(y_ t \vert \vy_{1:t-1})^{-1}\frac{p(\vy_{1:t-1})}{\hat{p}(\vy_{1:t-1})} \right].
\end{align}
As the number of samples goes to infinity~\eqref{eq:filtering_elbo} can be made arbitrarily tight; as a result, the \textit{implicit} approximation to the filtering distribution~\eqref{eq:implicit_approx} will become arbitrarily close to the true posterior, $p(x_t \vert \vy_{1:t})$, almost everywhere which allows for a trade-off between accuracy and computational complexity.
We note that this result is not applicable in most cases of VI due to the simplicity of variational families used.
We summarize this result in the following theorem (see the proof in section~\ref{prood:filtering_elbo} of the appendix).
\begin{theorem}[Filtering ELBO]
	\label{thm:filtering_elbo}
	The filtering ELBO \eqref{eq:filtering_elbo}, $\widetilde{\mathcal{L}}_t$, is a lower bound to the logarithm of the normalization constant of the filtering distribution, $p(x_t \vert \vy_{1:t})$. As the number of samples, $N$, goes to infinity, $\widetilde{\mathcal{L}}_t$ will become arbitrarily close to $\log p(y_t \vert \vy_{1:t-1})$.
\end{theorem}

Theorem \ref{thm:filtering_elbo} leads to the following corollary~\cite{del1996non} (proof in section~\ref{proof:corollary} of the appendix).
\begin{corollary}\label{corr:filtering_elbo}
	Theorem~\ref{thm:filtering_elbo} implies that the implicit filtering distribution, $\tilde{q}(x_t \vert \vy_{1:t})$, converges to the true posterior, $p(x_t~\vert~\vy_{1:t})$, as $N \to \infty$.
\end{corollary}

\subsection{Stochastic Optimization}
As in variational inference, we fit the parameters of the proposal, dynamics and observation model, $\Theta_t = \{\lambda_t, \theta_t, \psi_t\}$, by maximizing the (filtering) ELBO (Alg.~\ref{alg:smc}).
While the expectations in \eqref{eq:filtering_elbo} are not in closed form, we can obtain unbiased estimates of $\widetilde{\mathcal{L}}_t$ and its gradients with Monte Carlo.
Note that when obtaining gradients with respect to $\Theta_t$, we only need to take gradients of $\mathbb{E}[\log \hat{p}(y_t \vert \vy_{1:t-1})]$.
We also assume that the proposal distribution, $r(x_t \vert x_{t-1}, y_t;\lambda_t)$, is reparameterizable, i.e. we can sample from $r(x_t \vert x_{t-1}, y_t;\lambda_t)$ by setting $x_t = h(x_{t-1}, y_t, \epsilon_t; \lambda_t)$ for some function $h$ where $\epsilon_t \sim s(\epsilon_t)$ and $s$ is a distribution independent of $\lambda_t$.
Thus we can express the gradient of \eqref{eq:filtering_elbo} using the reparameterization trick~\cite{Kingma2014} as
\begin{equation}\label{eq:stochastic_gradient}
	\begin{aligned}
		\nabla_{\Theta_t} \widetilde{\mathcal{L}}_t
		 & = \nabla_{\Theta_t} \mathbb{E}_{s(\epsilon^{1:L})}
		[\log \hat{p}(y_t \vert \vy_{1:t-1})],
		\\
		 & = \mathbb{E}_{s(\epsilon^{1:L})}
		[\nabla_{\Theta_t} \log \hat{p}(y_t \vert \vy_{1:t-1})],
		\\
		 & = \mathbb{E}_{s(\epsilon^{1:L})}
		\left[
			\nabla_{\Theta_t}  \log \big( \sum_{i=1}^L  w_t^i \big)
			\right].
	\end{aligned}
\end{equation}
where $L \leq N$ is the number of subsamples to accelerate calculations.
In Algorithm~\ref{alg:smc}, we perform $N_{\text{SGD}}$ stochastic gradient descent (SGD) updates for each step.

While using more samples, $N$, will reduce the variational gap between the filtering ELBO, $\widetilde{\mathcal{L}}_t$, and $\log p(y_t \vert \vy_{1:t-1})$,
using more samples, $L$, for estimating \eqref{eq:stochastic_gradient} may be detrimental for optimizing the  parameters, as it has been shown to decrease the signal-to-noise ratio (SNR) of the gradient estimator for importance-sampling-based ELBOs~\cite{rainforth2018tighter}.
The intuition is as follows: as the number of samples used to compute the gradient increases, the bound gets tighter and tighter which in turn causes the magnitude of the gradient to become smaller. The rate at which the magnitude decreases is much faster than the variance of the estimator, thus driving the SNR to $0$.
In practice, we found that using a small number of samples to estimate ~\eqref{eq:stochastic_gradient}, $L < 5$, is enough to obtain good performance.

\subsection{Learning Dynamics with Sparse Gaussian Processes}\label{sec:proposal}
State-space models allow for various time series models to represent the evolution of state and ultimately predict the future~\cite{Shumway2010}.
While in some scenarios there exists prior knowledge on the functional form of the latent dynamics, $f_{\theta}(x)$, this is usually never the case in practice; thus $f_\theta(x)$ must be learned online as well.
While one can assume a parametric form for $f_\theta(x)$, i.e. a recurrent neural network, and learn $\theta$ through SGD, this does not allow uncertainty over the dynamics to be expressed which is key for many real-time, safety-critical tasks.
An attractive alternative over parametric models are Gaussian processes (GPs)~\cite{Rasmussen2006}.
Gaussian processes do not assume a functional form for the latent dynamics; rather, general assumptions, such as continuity or smoothness, are imposed.
Gaussian processes also allow for a principled notion of uncertainty, which is key when predicting the future.

A Gaussian process is defined as a collection of random variables, any finite number of which have a joint Gaussian distribution.
It is completely specified by its mean and covariance functions.
A GP allows one to specify a prior distribution over functions
\begin{equation}\label{eq:gp}
	f(x) \sim \mathcal{GP}(m(x), k(x, x'))
\end{equation}
where $m(\cdot)$ is the mean function and $k(\cdot, \cdot)$ is the covariance function; in this study, we assume that $m(x) = x$.
With the GP prior imposed on the dynamics, one can do Bayesian inference with data.

With the current formulation, a GP can be incorporated by augmenting the state-space to $(x_t, f_t)$, where $f_t \equiv f(x_{t-1})$ .
The importance weights are now computed according to
\begin{equation}\label{eq:naive_gp}
	w_t = \frac{p(y_t \vert x_t)p(x_t \vert f_t)p(f_t \vert \vf_{1:t-1}, \vx_{0:t-1})}{r(x_t, f_t \vert f_{t-1}, x_{t-1}, y_t; \lambda_t)}.
\end{equation}
Examining~\eqref{eq:naive_gp}, it is evident that naively using a GP is impractical for online learning because its space and time complexity are proportional to the number of data points, which grows with time $t$, i.e., $\mathcal{O}(t^2)$ and $\mathcal{O}(t^3)$ respectively.
In other words, the space and time costs increase as more and more observations are processed.

To overcome this limitation, we employ the sparse GP method~\cite{Titsias2009,Snelson2006}.
We introduce $M$ inducing points, $\vz = (z_1, \ldots, z_M)$, where $z_i = f(u_i)$ and $u_i$ are pseudo-inputs and impose that the prior distribution over the inducing points is $p(\vz) = \mathcal{N}(\mathbf{0}, k(\vu, \vu'))$.
In the experiments, we spread the inducing points uniformly over a finite volume in the latent space.
Under the sparse GP framework, we assume that $\vz$ is a sufficient statistic for $f_t$, i.e.
\begin{equation}\label{eq:non_marg_sgp}
	\begin{split}
		p(f_t & \vert \vx_{0:t-1}, \vf_{1:t-1}, \vz)
		= p(f_t \vert x_{t-1}, \vz) \\
		&= \mathcal{N}\left(f_t \vert m_t + K_{tz}K_{zz}^{-1}\vz, K_{tt} - K_{tz}K^{-1}_{zz}K_{zt}\right),
	\end{split}
\end{equation}
where $m_t = m(x_{t-1})$.
Note that the inclusion of the inducing points in the model reduces the computational complexity to be constant with respect to time.
Marginalizing out $f_t$ in~\eqref{eq:non_marg_sgp}
\begin{equation}\label{eq:sgp_dynamics}
	\begin{split}
		& p(x_t \vert x_{t-1}, \vz) \\
		&= \int p(x_t \vert f_t)p(f_t \vert x_{t-1}, \vz)df_t\\
		&= \mathcal{N}\left(x_t \vert m_t +  K_{tz}K_{zz}^{-1}\vz, K_{tt} - K_{tz}K^{-1}_{zz}K_{zt} + Q\right).
	\end{split}
\end{equation}
Equipped with equation~\eqref{eq:sgp_dynamics}, we can express the smoothing distribution as
\begin{equation}\label{eq:smoothing_sgp_no_diffusion}
	p(\vx_{0:t}, \vz \vert \vy_{1:t}) \propto p(x_0) p(\vz) \prod p(y_t \vert x_t)p(x_t \vert x_{t-1}, \vz),
\end{equation}
and the importance weights can be computed according to
\begin{equation}\label{eq:iw_non_marg_sgp}
	w_t = \frac{p(y_t \vert x_t)p(x_t \vert x_{t-1}, \vz)p(\vz \vert \vx_{0:t-1})}{r(x_t, \vz \vert x_{t-1}, y_t; \lambda_t)}.
\end{equation}
Due to the conjugacy of the model, $p(\vz \vert \vx_{0:t-1})$ can be recursively updated efficiently.
Let $p(\vz_t \vert \vx_{0:t-1}) = \mathcal{N}(\vz_t \vert \mu_{t-1}, \Gamma_{t-1})$.
Given $x_t$ and by Bayes rule
\begin{equation}
	p(\vz \vert \vx_{0:t}) \propto p(x_t \vert x_{t-1}, \vz) p(\vz \vert \vx_{0:t-1}),
\end{equation}
we obtain the recursive updating rule:
\begin{equation}\label{eq:recursive_updates_sgp}
	\begin{aligned}
		\Gamma_t & = \left(\Gamma_{t-1}^{-1}
		+ A_{t}^\top C_{t}^{-1} A_{t}\right)^{-1},
		\\
		\mu_t    & = \Gamma_t \left[
			\Gamma_{t-1}^{-1} \mu_{t-1} + A_{t}^\top C_{t}^{-1} (x_t - m_t)
			\right],
	\end{aligned}
\end{equation}
where $A_t = K_{tz} K_{zz}^{-1}$ and $C_t = K_{tt} - K_{tz} K_{zz}^{-1} K_{zt} + Q$.

To facilitate learning in non-stationary environments, we impose a diffusion process over the inducing variables.
Letting $p(\vz_{t-1} \vert x_{0:t-1}) = \mathcal{N}(\mu_{t-1}, \Gamma_{t-1})$, we impose the following relationship between $\vz_{t-1}$ and $\vz_t$
\begin{equation}
	\vz_t = \vz_{t-1} + \eta_t,
\end{equation}
where $\eta_t \sim \mathcal{N}(0, \sigma_z^2 I)$.
We can rewrite~\eqref{eq:iw_non_marg_sgp}
\begin{equation}\label{eq:iw_non_marg_sgp_diffusion}
	w_t = \frac{p(y_t \vert x_t)p(x_t \vert x_{t-1}, \vz_t)p(\vz_t \vert \vx_{0:t-1})}{r(x_t, \vz_t \vert x_{t-1}, \vz_{t-1}, y_t; \lambda_t)},
\end{equation}
where
\begin{equation}
	\begin{split}
		p(\vz_t \vert \vx_{0:t-1}) &= \int p(\vz_t \vert \vz_{t-1})p(\vz_{t-1} \vert \vx_{0:t-1}) d\vz_{t-1} \\
		&= \mathcal{N}(\mu_{t-1}, \Gamma_{t-1} + \sigma_z^2 I).
	\end{split}
\end{equation}
To lower the computation we marginalize out the inducing points from the model, simplifying~\eqref{eq:iw_non_marg_sgp_diffusion}
\begin{equation}\label{eq:iw_sgp_marg_diffusion}
	w_t = \frac{p(y_t \vert x_t)p(x_t \vert \vx_{0:t-1})}{r(x_t \vert x_{t-1}, y_t;\lambda_t)},
\end{equation}
where
\begin{equation}
	\begin{split}
		p(x_t \vert \vx_{0:t-1}) &= \int p(x_t \vert x_{t-1}, \vz_t)p(\vz_t \vert \vx_{0:t-1}) d\vz_t \\
		& = \mathcal{N}(v_t, \Sigma_t)
	\end{split}
\end{equation}
where $v_t = m_t + A_{t}\mu_{t-1}$ and $\Sigma_t = C_t + A_t(\Gamma_{t-1} + \sigma_x^2I)A_t^\top$.

For each stream of particles, we store $\mu_t^i$ and $\Gamma_t^i$.
Due to the recursive updates~\eqref{eq:recursive_updates_sgp}, maintaining and updating $\mu_t^i$ and $\Gamma_t^i$ is of constant time complexity, making it amenable for online learning.
The use of particles also allows for easy sampling of the predictive distribution (details are in section~\ref{sec:supp:prediction} of the appendix).
We call this approach SVMC-GP; the algorithm is summarized in Algorithm~\ref{alg:smc_gp}.
\noindent
\begin{algorithm}
	\caption[]{SVMC-GP (Step $t$)}
	\label{alg:smc_gp}
	\begin{algorithmic}[1]
		\Require $\{x_{t-1}^i, \mu_{t-1}^i, \Gamma_{t-1}^i, w_{t-1}^i\}_{i=1}^N, \Theta_{t-1}, y_t, \alpha$
		\For{$k = 1, \ldots, N_{\text{SGD}}$}
		\For{$i = 1, \ldots, L$}
		\State $a_t^i \sim \Pr(a_{t}^i = j) \propto w_{t-1}^j$ \Comment {\textit{Resample}}
		\State $x_t^i \sim r(x_t \vert y_t, x_{t-1}^{a_t^i}; \mu_{t-1}^{a_t^i}, \Gamma_{t-1}^{a_t^i}, \Theta_{t-1})$ \Comment {\textit{Propose}}
		\State $w_t^i \gets \frac{p(x_t^i \vert x_{t-1}^{a_t^i}) p(y_t \vert x_{t-1}^{a_t^i}; \Theta_{t-1})}{r(x_t^i \vert x_{t-1}^{a_t^i}, y_t; \mu_{t-1}^{a_t^i}, \Gamma_{t-1}^{a_t^i}, \Theta_{t-1})}$ \Comment {\textit{Reweigh}}
		\EndFor
		\State $\tilde{\mathcal{L}_t} \gets \log(\sum_{i}  w_t^i)$
		\State $\Theta_t \gets \Theta_{t-1} + \alpha \nabla_{\Theta} \tilde{\mathcal{L}_t}$ \Comment{SGD}
		\EndFor
		\State Resample, propose and reweigh $N$ particles
		\State Compute $\mu_t^i$ and $\Gamma_t^i$\\
		\Return $\{x_t^i, \mu_t^i, \Gamma_t^i, w_{t}^i\}_{i=1}^N$, $\Theta_t$
	\end{algorithmic}
\end{algorithm}

\subsection{Design of Proposals}
As stated previously, the accuracy of SVMC depends crucially on the functional form of the proposal.
The (locally) optimal proposal is
\begin{equation}\label{eq:optimal_proposal}
	p(x_t \vert x_{t-1}, y_t)  \propto p(y_t \vert x_t)  p(x_t \vert x_{t-1}),
\end{equation}
which is a function of $y_t$ and $f_t$~\cite{doucet2000sequential}.
In general~\eqref{eq:optimal_proposal} is intractable; to emulate~\eqref{eq:optimal_proposal} we parameterize the proposal as
\begin{equation}\label{eq:mlp_proposal}
	r(x_t \vert x_{t-1}, y_t) = \mathcal{N}(\mu_{\lambda_t}(f_t, y_t), \sigma^2_{\lambda_t}(f_t, y_t)I),
\end{equation}
where $\mu_{\lambda_t}$ and $\sigma_{\lambda_t}$ are neural networks with parameters $\lambda_t$.

\section{Related Works}
Much work has been done on learning good proposals for SMC.
The method proposed in \cite{guarniero2017iterated} iteratively adapts its proposal for an auxiliary particle filter.
In \cite{cornebise2014adaptive}, the proposal is learned online using expectation-maximization but the class of dynamics for which the approach is applicable for is extremely limited.
The method proposed in \cite{gu2015neural} learns the proposal by minimizing the KLD between the smoothing distribution and the proposal, $\KL[p(\vx_{0:t} \vert \vy_{1:t}) \Vert r(\vx_{0:t};\bm{ \lambda}_{0:t})]$; while this approach can be used to learn the proposal online, biased importance-weighted estimates of the gradient are used which can suffer from high variance if the initial proposal is bad.
Conversely, \cite{Naesseth2018} maximizes $\mathbb{E}[\log \hat{p}(\vy_{1:t})]$, which can be shown to minimize the KLD between the proposal and the implicit smoothing distribution, $\KL[q(\vx_{0:t} \vert \vy_{1:t}) \Vert p(\vx_{0:t} \vert \vy_{1:t})]$; biased gradients were used to lower the variance.
In contrast, SVMC allows for \textit{unbiased} and \textit{low variance} gradients that target the filtering distribution as opposed to the smoothing distribution.
In~\cite{xu2019learning}, the proposal is parameterized by a Riemannian Hamiltonian Monte Carlo and the algorithm updates the mass matrix by maximizing $\mathbb{E}[\log \hat{p}(\vy_{1:t})]$.
At each time step (and for every stochastic gradient), the Hamiltonian must be computed forward in time using numerical integration, making the method impractical for an online setting.

Previous works have also tackled the dual problem of filtering and learning the parameters of the model online. 
A classic approach is to let the parameters of the generative model evolve according to a diffusion process, $\theta_t = \theta_{t-1} + \upsilon_t$: one can then create an augmented latent state, $\tilde{x}_t = [x_t, \theta_t]$, and perform filtering over $\tilde{x}_t$ either using particle filtering~\cite{liu2001combined} or joint extended/unscented Kalman filtering~\cite{wan1997dual,wan1999dual}.
One can also use a separate filter for the latent state and the parameters, which is known as dual filtering~\cite{wan1997dual,wan1999dual}.
As SVMC is a general framework, we could also let the parameters of the generative model evolve according to a diffusion process and do joint/dual filtering; the availability of the filtering ELBO allows us to learn the variance of the diffusion online, while previous approaches treat this a fixed hyper-parameter.
Besides, as we demonstrate in later experiments, we can learn the parameters of a parametric model online by performing SGD on the filtering ELBO.
In~\cite{deisenroth2011robust}, they combine extended Kalman filtering (EKF) with Gaussian processes for dual estimation; the use of EKF involves approximations and restricts the observation models one can apply it on. Moreover, the use of a full Gaussian process--as opposed to a sparse alternative--prevents it from being deployed for long time series.
In~\cite{Ko2009}, particle filtering is combined with a sparse Gaussian process to learn the latent dynamics and the emission model online; while similar to SVMC-GP, there are important differences between the two works.
Firstly--and most importantly--the latent fixed dynamics are \textbf{not} learned online in~\cite{Ko2009}; training data is collected a priori and used to pre-train the GP and is kept during the filtering process.
While a priori training data can also be used for SVMC-GP, our formulation allows us to continuously learn the latent dynamics in a purely online fashion.
Second, a fixed proposal--similar to the one found in bootstrap particle filtering--is used while SVMC-GP allows for the proposal to adapt on-the-fly. 
In~\cite{campbell2021online}, they tackle the problem of dual estimation by leveraging the recursive form of the smoothing distribution to obtain an ELBO that can be easily computed online, allowing for the parameters of the generative model to be inferred using SGD.
While similar to SVMC, we note that their approach relies on simple parametric variational approximations which are not as expressive as the particle based ones used in SVMC.

\begin{figure}[thb]
	\includegraphics[width=\linewidth]{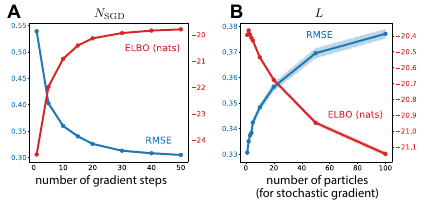}
	\caption{Investigating how the performance of SVMC--measured via RMSE (lower is better) and ELBO (higher is better)--depends on number of gradient steps, $N_{\text{SGD}}$ and number of particles used to compute stochastic gradient, $L$.
	For each setting, we run 100 realizations of SVMC on the chaotic RNN data from sec. 4.1.2.
	Solids lines are the average where error bars are the standard error.
		A) For a fixed number of particles used to compute stochastic gradient, $L=4$, the number of gradient steps, $N_{\text{SGD}}$ taken at every time step is varied.
	B) For a fixed number of gradient steps, $N_{\text{SGD}} = 15$, the number of particles used to compute stochastic gradient, $L$, is varied.
		}
	\label{fig:convergence}
\end{figure}

\begin{figure}[t!h]
	\centering
	\includegraphics[width=\linewidth]{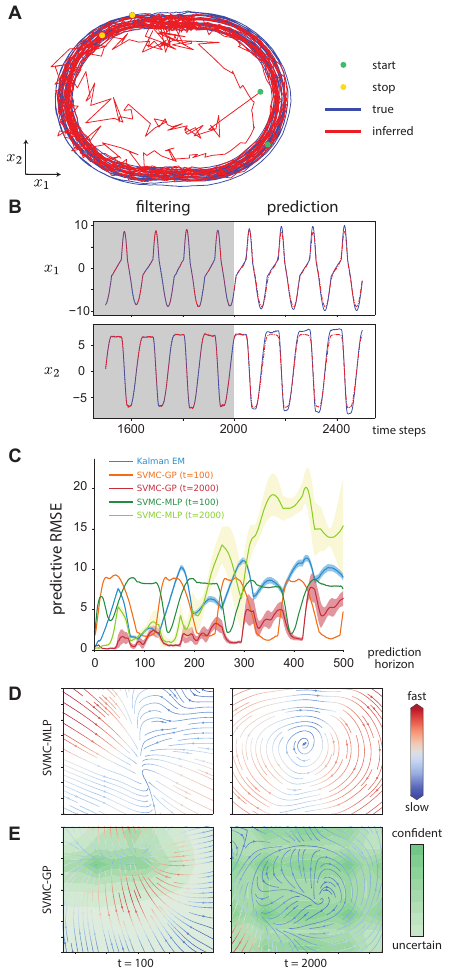}
	\caption[NASCAR]{NASCAR\textsuperscript{\textregistered} Dynamics~\cite{linderman2017bayesian}.
		(A) \textcolor{blue}{True} and inferred \textcolor{red}{latent} trajectory using SVMC-GP.
		(B) Filtering and prediction. We show the last $500$ steps of filtered states and the following $500$ steps of predicted states.
		(C) Forecasting error. We compare the $500$-step predictive MSE (averaged over $100$ realizations) of SVMC-GP, SVMC-MLP, and Kalman filter.  The transition matrix of the Kalman filter was learned by EM (offline). The periodic tendency is due to the periodic nature of ground truth.
		(D)--(E) Inferred dynamics as velocity field.
		For SVMC-GP, posterior variance of dynamics is additionally shown as uncertainty.
	}
	\label{fig:nascar}
\end{figure}

\section{Experiments}
To showcase the power of SVMC, we employ it on a number of simulated and real experiments.
For all experiments, the Adam optimizer was used~\cite{kingma2015}.
\subsection{Synthetic Data}
\subsubsection{Linear Dynamical System}
As a baseline, we apply SVMC on data generated 
from a linear dynamical system (LDS)
\begin{equation}\label{eq:LDS}
	\begin{aligned}
		x_t & = Ax_{t-1} + \epsilon_t, & \epsilon_t \sim \mathcal{N}(0, Q), \\
		y_t & = Cx_t + \xi_t,          & \xi_t \sim \mathcal{N}(0, R).
	\end{aligned}
\end{equation}
LDS is the \textit{de facto} dynamical system for many fields of science and engineering due to its simplicity and efficient exact filtering (i.e., Kalman filtering).
The use of an LDS also endows a closed form expression of the log marginal likelihood for the filtering and smoothing distribution.
Thus, as a baseline we compare the estimates of the negative log marginal likelihood, $-\log p(\vy_{1:T})$, produced by SVMC, variational sequential Monte Carlo (VSMC)~\cite{Naesseth2018} (which is an offline method) and BPF~\cite{gordon1993novel} in an experiment similar to the one used in~\cite{Naesseth2018}.
We generated data from \eqref{eq:LDS} with $T=50$, $d_x=d_y=10$, $(A)_{ij} = \alpha^{\vert i - j \vert + 1}$, with $\alpha=0.42$ and $Q=R=I$ where the state and emission parameters are fixed; the true negative log marginal likelihood is $1168.2$.
For SVMC and VSMC, we used the same proposal parameterization as~\cite{Naesseth2018}
\begin{equation}\label{eq:lds_proposal}
	r(x_t \vert x_{t-1}; \lambda_t) = \mathcal{N}(\mu_t + \textrm{diag}(\beta_t) Ax_{t-1} , \textrm{diag}(\sigma_t^2)),
\end{equation}
where $\lambda_t = \{\mu_t, \beta_t, \sigma_t^2\}$.
To ensure VSMC has enough time to converge, we use 25,000 gradient steps.
To equate the total number of gradient steps between VSMC and SVMC, 25,000/50 = 500 gradient steps were done at each time step for SVMC.
For both methods, a learning rate of 0.01 was used where $L=4$ particles were used for computing gradients, which was used in~\cite{Naesseth2018}.
To equate the computational complexity between SVMC and BPF, we ran the BPF with 125,000 particles.
We fixed the data generated from~\eqref{eq:LDS} and ran each method for 100 realizations; the average negative ELBO and its standard error of each the methods are shown in Table~\ref{tab:lds}.
To investigate the dependence of the ELBO on the number of particles, we demonstrate results for SVMC and VSMC using a varying number of particles.

From Table~\eqref{tab:lds}, we see that SVMC performs better than VSMC for all number of particles considered.
While SVMC with 100 particles is outperformed by BPF, SVMC with 1,000 particles matches the performance of BPF with a smaller computational budget.
\begin{table*}[t]
	\centering
	\small
	\caption{Experiment 1 (LDS) with 100 replication runs (true negative log-likelihood is $1168.12$).
		The average negative ELBO and runtime are shown with the standard error for SVMC, VSMC and BPF where the number in parenthesis is the number of particles used.}
	\label{tab:lds}
	\begin{tabular}{
			c
			|
			S[table-format=4.1,detect-weight,mode=text]
			@{${}\pm{}$}
			S[table-format=2.1,detect-weight,mode=text]
			S[table-format=4.1,detect-weight,mode=text]
			@{${}\pm{}$}
			S[table-format=2.1,detect-weight,mode=text]
			S[table-format=4.1,detect-weight,mode=text]
			@{${}\pm{}$}
			S[table-format=2.1,detect-weight,mode=text]
			S[table-format=4.1,detect-weight,mode=text]
			@{${}\pm{}$}
			S[table-format=2.1,detect-weight,mode=text]
			S[table-format=4.1,detect-weight,mode=text]
			@{${}\pm{}$}
			S[table-format=2.1,detect-weight,mode=text]
			S[table-format=4.1,detect-weight,mode=text]
			@{${}\pm{}$}
			S[table-format=2.1,detect-weight,mode=text]
			S[table-format=4.1,detect-weight,mode=text]
			@{${}\pm{}$}
			S[table-format=2.1,detect-weight,mode=text]
			S[table-format=4.1,detect-weight,mode=text]
			@{${}\pm{}$}
			S[table-format=2.1,detect-weight,mode=text]
		}
		 & \multicolumn{2}{c}{SVMC (100)}
		 & \multicolumn{2}{c}{VSMC (100)}
		 & \multicolumn{2}{c}{SVMC (1,000)}
		 & \multicolumn{2}{c}{VSMC (1,000)}
		 & \multicolumn{2}{c}{SVMC (10,000)}
		 & \multicolumn{2}{c}{VSMC (10,000)}
		 & \multicolumn{2}{c}{BPF (125,000)}
		\\\hline
		$-$ELBO
		 & 1188.3                            & 0.5
		 & 1195.9                            & 0.5
		 &  1178.3                            &  0.3
		 & 1183.6                            & 0.3
		 & \bfseries 1173.8                            & 0.2
		 & 1179.8                            & 0.2
		 & 1177.0                            & 0.2
		\\
		time (s)
		 & 47.5                             & 0.5
		 & 6390.2                             & 3.1
		 & 51.6                            & 0.3
		 & 6390.2                             & 3.1
		 & 64.5                              & 0.5
		 & 6390.2                              & 3.1
		 & 95.0                              & 0.7
	\end{tabular}
\end{table*}

\subsubsection{Chaotic Recurrent Neural Network}
To show the performance of our algorithm in filtering data generated from a complex, nonlinear and high-dimensional dynamical system, we generate data from a continuous-time "vanilla" recurrent neural network (vRNN)
\begin{equation}\label{eq:cont_time_vrnn}
	\tau \dot{x}(t)=-x(t)+\gamma W_x \tanh (x(t)) + \sigma(x)dW(t).
\end{equation}
where $W(t)$ is Brownian motion.
Using Euler integration, \eqref{eq:cont_time_vrnn} can be described as a discrete time dynamical system
\begin{equation}\label{eq:discrete_time_vrnn}
	x_{t+1} = x_t + \Delta ( -x_t + \gamma W_x \tanh (x_t))/ \tau + \epsilon_t, \quad \epsilon_t \sim \mathcal{N}(0, Q)
\end{equation}
where $\Delta$ is the Euler step.
The emission model is
\begin{equation}\label{eq:vrnn_student_t}
	y_t = Cx_t + D + \xi_t, \quad \xi_{t, 1}, \cdots, \xi_{t, d_y} \stackrel{\textrm{i.i.d}}{\sim} \mathcal{ST}(0, \nu_y, \sigma_y)
\end{equation}
where each dimension of the emission noise, $v_t$, is independently and identically distributed (i.i.d) from a Student's t distribution, $\mathcal{ST}(0, \nu_y, \sigma_y)$, where $\nu_y$ is the degrees of freedom and $\sigma_y$ is the scale.

\begin{figure}[t!h!b!]
    \centering
    \includegraphics[width=0.97\linewidth]{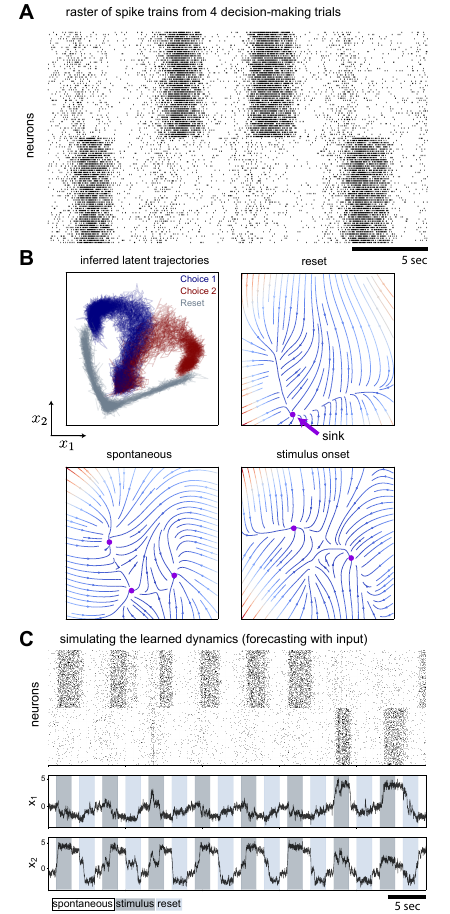}
    \caption[Winner-Take-All Spiking Neural Network]{Winner-Take-All Spiking Neural Network.
        (A) 4 trials of training data. The neuronal activity was drawn over a 25 sec time window. Each row represents one neuron. Each dot represents that neuron fired within that time bin.
        (B) Inference. The top-left panel shows the inferred latent trajectories of several trials. In each trial the network started at the initial state, eventually reached either choice (indicated by the color) after the stimulus onset, and finally went back around the initial state after receiving reset signal.
        The rest three panels show the phase portraits of inferred dynamical system revealing the bifurcation and how the network dynamics were governed at different phases of the experiment. 
        At the spontaneous phase (when the network receive no external input), the latent state is attracted by the middle sink. After the stimulus is onset, the middle sink disappears and the latent state falls into either side driven by noise to form a choice.
        When the reset is onset, the latent state is pushed back to the only sink that is close to the middle sink of the spontaneous phase, and then the network is ready for a new trial.
        (C) Simulation from the fitted model. We generated latent trajectory and spike train by replicating the experiments on the fitted model. The result shows that the model can replicate the dynamics of the target network. 
    }
    \label{fig:wang}
\end{figure}

We set $d_y = d_x = 10$ and the elements of $W_x$ are i.i.d. drawn from $\mathcal{N}(0, 1/d_x)$. We set $\gamma = 2.5$ which produces chaotic dynamics at a relatively slow timescale compared to $\tau$~\cite{SUSSILLO2009544}.
The rest of the parameters values are: $\tau=0.025$, $\delta=0.001$, $Q = 0.01I$, $\nu_y=2$ and $\sigma_y = 0.1$, which are kept fixed.
We generated a single time series of length of $T=500$ and fixed it across multiple realizations.
SVMC was ran using 200 particles with proposal distribution~\eqref{eq:mlp_proposal}, where the neural network was a multi-layer perceptron (MLP) with 1 hidden layer of width 100 and relu nonlinearities; 15 gradient steps were performed at each time step with a learning rate of .001 with $L=4$.
For a comparison, a BPF with 10,000  particles and an unscented Kalman filter (UKF) was run.
Each method was ran over 100 realizations.
We compare the ELBO and root mean square error (RMSE) between the true latent states, $x_{1:T}$, and the inferred latent states, $\hat{x}_{1:T}$.\footnote{We use the posterior mean as our estimate of the latent states.} 
\begin{table}[ht]
	\small
	\centering
	\caption{Experiment 2 (Chaotic RNN) with 100 replication runs.
		The average RMSE (lower is better), negative ELBO (lower is better) and runtime per step are shown with standard error.}
	\begin{tabular}{
			c
			|
			S[table-format=2.2,detect-weight,mode=text]
			@{${}\pm{}$}
			S[table-format=1.2,detect-weight,mode=text]
			S[table-format=2.2,detect-weight,mode=text]
			@{${}\pm{}$}
			S[table-format=1.2,detect-weight,mode=text]
			S[table-format=2.2,detect-weight,mode=text]
			@{${}\pm{}$}
			S[table-format=1.2,detect-weight,mode=text]
			S[table-format=2.1,detect-weight,mode=text]
			@{${}\pm{}$}
			S[table-format=1.2,detect-weight,mode=text]
		}
		 & \multicolumn{2}{c}{SVMC (200)}
		 & \multicolumn{2}{c}{BPF (10,000)}
		 & \multicolumn{2}{c}{UKF}
		\\\hline
		RMSE
		 & \bfseries .34                    & \bfseries .001
		 & .4                             & .002
		 & 3.9                              & .12
		\\
		$-$ELBO (nats)
		 & \bfseries 20.42                   & \bfseries .008
		 & 24.16                             & .018
		 & \multicolumn{2}{c}{N/A}
		\\
		time (s)
		 & 18.78                             & .08           
		 & 15.83                             & .09
		 & 0.8                              & .004
	\end{tabular}
	\label{tab:exp2}
\end{table}
With a similar computational budget, SVMC can achieve better performance than a BPF using almost two orders of magnitude less samples.
To investigate the effect the number of gradient steps has on the performance of SVMC, we plot the RMSE and ELBO as a function of number of gradient steps taken in figure~\ref{fig:convergence}A; taking more gradient steps leads to a decrease in RMSE and an increase in the ELBO.
We next investigate the effect the number of samples used to compute the stochastic gradient, $L$, has on the performance~\ref{fig:convergence}B; as was demonstrated in~\cite{rainforth2018tighter}, larger L leads to a decrease in performance.

\subsubsection{Synthetic NASCAR\textsuperscript{\textregistered} Dynamics}
We test learning dynamics online with sparse GP on a synthetic data of which the underlying dynamics follow a recurrent switching linear dynamical systems~\cite{linderman2017bayesian}. The simulated trajectory resembles the NASCAR\textsuperscript{\textregistered} track (Fig.~\ref{fig:nascar}A).
We train the model with $2,000$ observations simulated from
$y_t = C x_t + \xi_t$
where $C$ is a $50$-by-$2$ matrix.
The proposal is defined as $\mathcal{N}(\mu_t, \Sigma_t)$ of which $\mu_t$ and $\Sigma_t$ are linear maps of the concatenation of observation $y_t$ and previous state $x_{t-1}^i$.
We use 50 particles, squared exponential (SE) kernel and 20 inducing points for GP and 1E-4 learning rate. We also run a SVMC (with the same setting on particles and learning rate as the former) with MLP (1 hidden layer and 20 hidden units) dynamics for comparison. GP dynamics not only estimate the velocity field but also give the uncertainty over the estimate while MLP dynamics is only a point estimate.
To investigate the learning of dynamics, we control for other factors, i.e. we fix the observation model and other hyper-parameters such as noise variances at their true values. (See the details in section~\ref{sec:supp:nascar} of the appendix.)

Figure~\ref{fig:nascar}A shows the true (blue) and inferred (red) latent states.
The inference quickly catches up with the true state and stays on the track.
As the state oscillates on the track, the sparse GP learns a reasonable limit cycle (Fig.~\ref{fig:nascar}F) without seeing much data outside the track.
The velocity fields in Figure~\ref{fig:nascar}D--F show the gradual improvement in the online learning.
The $500$-step prediction also shows that the GP captures the dynamics (Fig.~\ref{fig:nascar}B).
We compare SVMC with Kalman filter in terms of mean squared error (MSE) (Fig.~\ref{fig:nascar}C).
The transition matrix of the LDS of the Kalman filter (KF) is learned by expectation-maximization which is an offline method, hence not truly online.
Yet, SVMC performs better than KF after $1000$ steps.

\subsubsection{Winner-Take-All Spiking Neural Network}
\begin{figure}[!ht]
	\centering
	\includegraphics[width=0.95\linewidth]{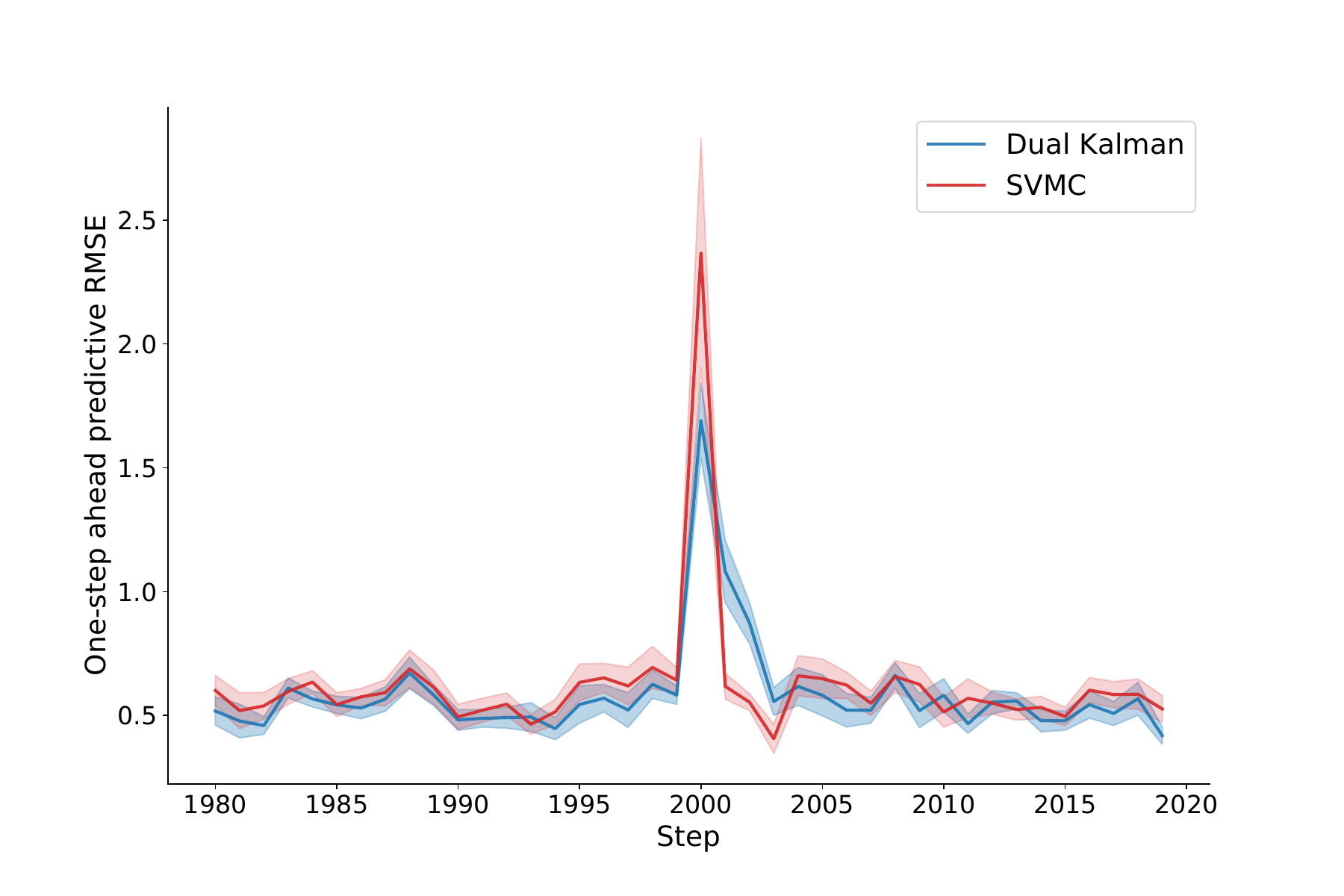}
	\caption{
		Prediction of nonstationary dynamical system. The colored curves (blue: EKF, red: SVMC) are the RMSEs (solid line: mean, shade: stderr) of one-step-ahead prediction of nonstationary system during online learning (50 trials each run, 10 runs). The linear system was changed and the state was perturbed at the 2000th step (center). Both online algorithms quickly learned the change after a few steps.
	}\label{fig:nonstationary}
\end{figure}

\begin{figure*}[!b!th]
	\centering
	\includegraphics[width=\linewidth]{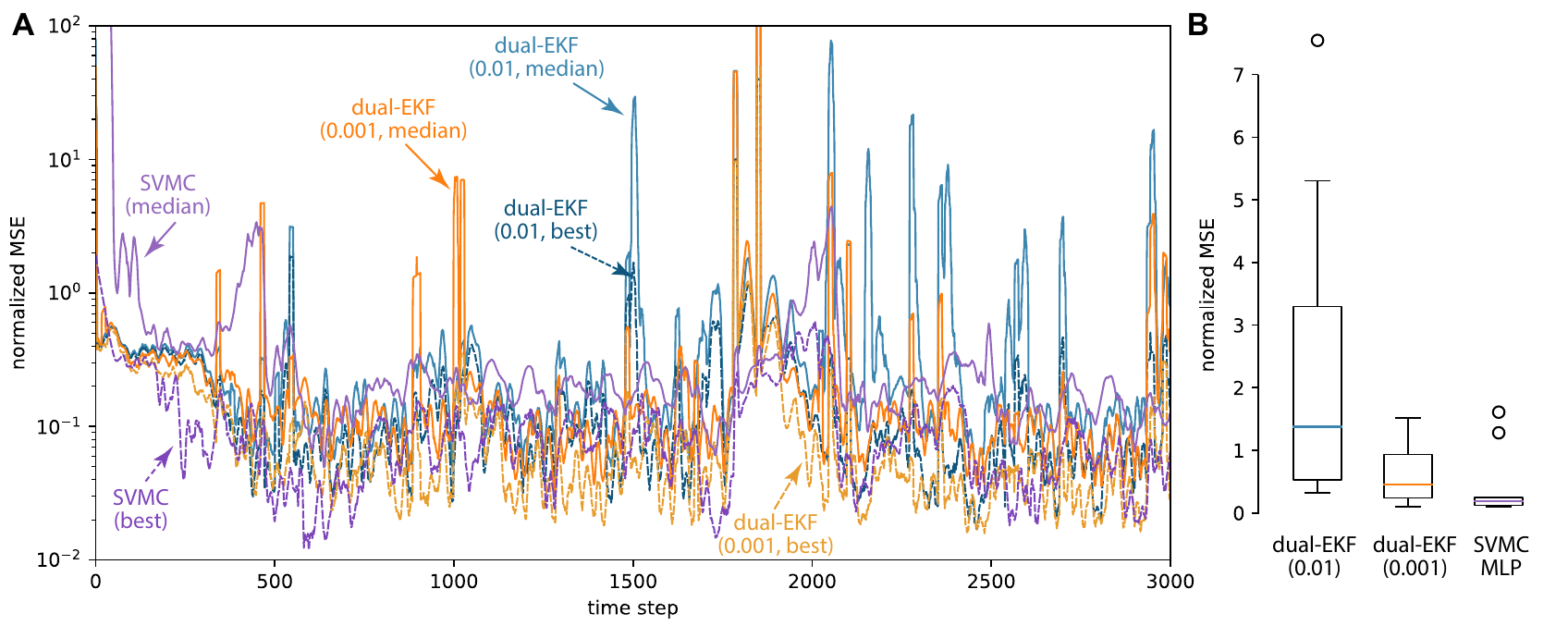}
	\caption{Prediction performance on 3D data generated from an analog stable oscillator circuit.
	    We compare Dual-EKF and SVMC both with dynamics parameterized with MLP (2-20-2).
	    (A) Normalized MSE of 100 time step prediction using the filtered system.
	    Median and best out of 11 randomly initialized filters are shown.
	    To estimate the normalized MSE, 11 realizations were used, and for ease of visual parsing 11 bin moving window averaging was applied.
	    (B) Comparison of normalized MSE of the last 500 time steps.
	}
	\label{fig:circuit}
\end{figure*}

The perceptual decision-making paradigm is a well-established cognitive task where typically a low-dimensional decision variable needs to be integrated over time, and subjects are close to optimal in their performance.
To understand how the brain implements such neural computation, many competing theories have been proposed~\cite{Barak2013,Mante2013,Ganguli2008,Wang2002,Wong2006}.
We test our method on a simulated biophysically realistic cortical network model for a visual discrimination experiment~\cite{Wang2002}.
In the model, there are two excitatory sub-populations that are wired with slow recurrent excitation and feedback inhibition to produce attractor dynamics with two stable fixed points.
Each fixed point represents the final perceptual decision, and the network dynamics amplify the difference between conflicting inputs and eventually generates a binary choice.

The simulated data was organized into decision-making trials.
We modified the network by injecting a $60$~Hz Poisson input into the inhibitory sub-population at the end of each trial to "reset'' the network for the purpose of uninterrupted trials to fit the streaming case because the original network was designed for only one-time experiment. In each trial the input to the network consisted of two periods, one 2-sec stimulus signal with different strength of visual evidence controlled by "coherence'', and one 2-sec $60$~Hz reset signal, each follows a 1-sec spontaneous period (no input). We subsampled 480 selective neurons out of 1600 excitatory neurons from the simulation to be observed by our algorithm.

Fig.~\ref{fig:wang} shows that SVMC ($300$ particles) with sparse GP dynamics ($150$ inducing points, squared exponential kernel) and MLP proposal (1 hidden layer, $1000$ hidden units) with $L=2$, learning rate 1E-4 and 15 gradient steps, did well at learning the dynamics of target network. The inferred latent trajectories of several trials (Fig.~\ref{fig:wang}B). In each trial the network started at the initial state, eventually reached either choice (indicated by the color) after the stimulus onset, and finally went back around the initial state after receiving reset signal.
The other three panels (Fig.~\ref{fig:wang}B) show the phase portraits of the inferred dynamical system revealing the bifurcation and how the network dynamics are governed during different phases of the experiment. 
In the spontaneous phase (when the network receives no external input), the latent state is attracted by the middle sink. After the stimulus is onset, the middle sink disappears and the latent state falls into either side driven by noise to form a choice.
When the reset is onset, the latent state is pushed back to the only sink that is close to the middle sink of the spontaneous phase, and then the network is ready for a new trial.
We generated a latent trajectory and corresponding spike train by replicating the experiments on the fitted model (Fig.~\ref{fig:wang}C). The result shows that the model can replicate the dynamics of the target network. 

The mean-field reduction of the network (Fig.~\ref{fig:wang_mean_field}~\cite{Wong2006}) also confirms that our inference accurately captured the key features of the network.
 Note that our inference was done without knowing the true underlying dynamics which means our method can recover the dynamics from data as a bottom-up approach.

\subsubsection{Nonstationary system}
Another feature of our method is that its state dynamics estimate never stops.
As a result, the algorithm is adaptive, and can potentially track slowly varying (nonstationary) latent dynamics.
To test adaptation to perturbation to both the state and system, we compared a dual EKF and the proposed approach (50 particles, GP dynamics with $10$ inducing points and squared exponential kernel, linear proposal, 1E-4 learning rate) on a 2D nonstationary linear dynamical system. A spiral-in linear system was suddenly changed from clockwise to counter-clockwise at the 2000th step and the latent state was perturbed (Fig.~\ref{fig:nonstationary}).
To adapt EKF, we used Gaussian observations that were generated through linear map from 2-dimensional state to 200-dimensional observation with additive noise ($\mathcal{N}(0, 0.5)$). To focus on the dynamics, we fixed all the parameters except the transition matrix for both methods, while our approach still has to learn the recognition model in addition. Our method quickly adapts in a few steps.

\subsection{Real Data: Learning an Analog Circuit}
It has been verified that the proposed methodology is capable of learning the underlying dynamics from noisy streaming observations in the above synthetic experiments. To test it in real world, we applied our method to the voltage readout from an analog circuit~\cite{Jordan2020}. 
We designed and built this circuit to realize a system of ordinary differential equations as follows
\begin{equation}
	\begin{aligned}
		\dot{x} & = (5z - 5) [x - \tanh(\alpha x - \beta y)] \\
		\dot{y} & = (5z - 5) [y - \tanh(\beta x + \alpha y)] \\
		\dot{z} & = -0.5 (z - \tanh(1.5z))
	\end{aligned}
\end{equation}
where $\cdot$ indicates the time derivative and $\alpha = \beta = 1.5\cos(\frac{\pi}{5})$.
This circuit performed oscillation with a period of approximately $2$~Hz. The sampling rate was $2000$~Hz.

We assume the following model:
\begin{align}
& x_t = f(x_{t-1}) + \epsilon_t, \\
& y_t = Cx + d + \psi_t,
\end{align}
where $x_t \in \mathbb{R}^2$, $y_t \in \mathbb{R}^3$, $\epsilon_t \sim \mathcal{N}(0, 10^{-3})$ and $\xi_t \sim \mathcal{N}(0, 10^{-3})$.
We parameterize $f(\cdot)$ using an MLP (1 hidden layer, 20 hidden units) and perform dual estimation using SVMC and dual EKF
on 3,000 time steps (Fig.~\ref{fig:circuit}A). We chose two different levels of diffusion ($0.001, 0.0001$) on the parameters for dual EKF to implement different learning rates. We forecast 10 realizations of 100 steps ahead every filtering step and show the mean and standard deviation of the logarithm of MSE to the true observation (Fig.~). 
As dual EKF has trouble learning the parameters of the observation model, we fixed $C$ and $d$ for dual EKF while we let SVMC (500 particles, lr 1E-4 and 15 gradient steps) learn both the parameters of the latent dynamics, $C$ and $d$.
Figure~ shows SVMC achieve the same level of performance but of less variance, and the slow convergence in the beginning was due to learning more parameters.
The inferred dynamics shows that the limit cycle can implement the oscillation (Fig.~\ref{fig:circuit}B). The prediction of future observations (500 steps) resemble the oscillation and the true observation is covered by 100 repeated predictions (Fig.~\ref{fig:circuit}C). The predictions started at the final state of the training data, and we simulated the future observation trajectory from the trained model without seeing any new data. We repeated the procedure of prediction 100 times. Figure.~\ref{fig:circuit}D shows the normalized predictive MSE (relative to the mean observation over time). The solid line is the mean normalized MSE and the shade is the standard error. Since the simulation included the state noise, the prediction diverged from the true observations as time goes.

\section{Discussion}
In this study we developed a novel online learning framework, leveraging variational inference and sequential Monte Carlo, which enables flexible and accurate Bayesian joint filtering.
Our derivation shows that our filtering posterior can be made arbitrarily close to the true one for a wide class of dynamics models and observation models.
Specifically, the proposed framework can efficiently infer a posterior over the dynamics using sparse Gaussian processes by augmenting the state with the inducing variables that follow a diffusion process. Taking benefit from Bayes' rule, our recursive proposal on the inducing variables does not require optimization with gradients.
Constant time complexity per sample makes our approach amenable to online learning scenarios and suitable for real-time applications.
In contrast to previous works, we demonstrate our approach is able to accurately filter the latent states for linear / nonlinear dynamics, recover complex posteriors, faithfully infer dynamics, and provide long-term predictions.
In future, we want to focus on reducing the computation time per sample that could allow for real-time application on faster systems. On the side of GP, we would like to investigate the priors and kernels that characterize the properties of dynamical systems as well as the hyperparameters.

\appendices
\section{Proof that $\hat{p}(y_t \vert \vy_{1:t-1})$ is a consistent estimator for $p(y_t \vert \vy_{1:t-1})$}\label{proof:consistent}
\begin{proof}
	To prove that $\hat{p}(y_t \vert \vy_{1:t-1})$ is a consistent estimator, we will rely on the delta method~\cite{van2000asymptotic}. From~\cite{chopin2004central}, we know that the central limit theorem (CLT) holds for $\hat{p}(\vy_{1:t})$ and $\hat{p}(\vy_{1:t-1})$
	\begin{gather}
		\sqrt{N}(\hat{p}(\vy_{1:t-1}) - p(\vy_{1:t-1})) \stackrel{d}{\to} \mathcal{N}(0, \sigma_{t-1}^2), \\
		\sqrt{N}(\hat{p}(\vy_{1:t}) - p(\vy_{1:t})) \stackrel{d}{\to} \mathcal{N}(0, \sigma_t^2)
	\end{gather}
	where we assume that $\sigma_{t-1}^2$ and $\sigma_t^2 $ are finite.
	We can express $\hat{p}(y_t \vert \vy_{1:t-1})$ as a function of $\hat{p}(\vy_{1:t})$ and $\hat{p}(\vy_{1:t-1})$,
	\begin{equation}
		\hat{p}(y_t \vert \vy_{1:t-1}) =g(\hat{p}(\vy_{1:t}), \hat{p}(\vy_{1:t-1})) =  \frac{\hat{p}(\vy_{1:t})}{\hat{p}(\vy_{1:t-1})}.
	\end{equation}
	Since $\frac{p(\vy_{1:t})}{p(\vy_{1:t-1})} = p(y_t \vert \vy_{1:t-1})$ and $g$ is  a continuous function, an application of the Delta method gives
	\begin{equation}
		\sqrt{N} (\hat{p}(y_t \vert \vy_{1:t-1}) - p(y_t \vert \vy_{1:t-1})) \stackrel{d}{\to} \mathcal{N}(0, \nabla g^\top \Sigma  \nabla g),
	\end{equation}
	where $\Sigma_{1,1} = \sigma_t^2$, $\Sigma_{2,2} = \sigma_{t-1}^2$ and $\Sigma_{1, 2} = \Sigma_{2,1} = \sigma_{t, t-1}$ where by the Cauchy-Schwartz inequality, $\sigma_{t, t-1}$ is also finite~\cite{van2000asymptotic}.
	Thus, as $N \rightarrow \infty$, $\hat{p}(y_t \vert \vy_{1:t-1})$ will converge in probability to $p(y_t \vert \vy_{1:t-1})$, proving the consistency of the estimator.
\end{proof}

\section{Proof of Theorem \ref{thm:filtering_elbo}}\label{prood:filtering_elbo}
\begin{proof}
	It is well known that the importance weights produced in a run of SMC are an unbiased estimator of $p(\vy_{1:t})$~\cite{Doucet2013}
	\begin{equation}
		\mathbb{E}[\hat{p}(\vy_{1:t})] = p(\vy_{1:t})
	\end{equation}
	where $\hat{p}(\vy_{1:t}) = \prod_{j=1}^t \frac{1}{N}\sum_{i=1}^N w_j^i$. We can apply Jensen's inequality to obtain
	\begin{equation}\label{eq:inequality}
		\log p(\vy_{1:t}) \geq \mathbb{E}[\log\hat{p}(\vy_{1:t})] .
	\end{equation}
	Expanding both sides of \eqref{eq:inequality}
	\begin{equation}
		\begin{split}
			& \log p(y_{t} \vert \vy_{1:t-1}) + \log p(\vy_{1:t-1}) \\
			& \qquad \geq \mathbb{E}[\log \hat{p}(y_t \vert \vy_{1:t-1})] + \mathbb{E}[\log \hat{p}(\vy_{1:t-1})].
		\end{split}
	\end{equation}
	Subtracting $\log p(\vy_{1:t-1})$ from both sides gives
	\begin{equation}
		\begin{split}
			& \log p(y_{t} \vert \vy_{1:t-1}) \\
			& \qquad \geq \mathbb{E}[\log \hat{p}(y_t \vert \vy_{1:t-1})] + \mathbb{E}[\log \hat{p}(\vy_{1:t-1})] - \log p(\vy_{1:t-1}).
		\end{split}
	\end{equation}
	Letting $\mathcal{R}_t(N) = \log p(\vy_{1:t-1}) -  \mathbb{E}[\log \hat{p}(\vy_{1:t-1})]$, where $N$ is the number of samples, we get
	\begin{equation}
		\log p(y_{t} \vert \vy_{1:t-1}) \geq \mathbb{E}[\log \hat{p}(y_t \vert \vy_{1:t-1})] - \mathcal{R}_t(N),
	\end{equation}
	where by Jensen's inequality~\eqref{eq:inequality}, $\mathcal{R}_t(N) \geq 0$ for all values of $N$. By the continuous mapping theorem~\cite{van2000asymptotic},
	\begin{equation}
		\lim_{N \to \infty } \mathbb{E}[ \log \hat{p}(\vy_{1:t-1})] = \log p(\vy_{1:t-1}).
	\end{equation}
	As a consequence, $\lim_{N \to \infty } \mathbb{E}[\mathcal{R}_t(N)] = 0$. By the same logic, and leveraging that $\hat{p}(y_t \vert \vy_{1:t-1})$ is a consistent estimator for $p(y_t \vert \vy_{1:t-1})$, we get that
	\begin{equation}
		\lim_{N \to \infty } \mathbb{E}[ \log \hat{p}(y_t \vert \vy_{1:t-1})] = \log p(y_t \vert \vy_{1:t-1}).
	\end{equation}
	Thus $\mathcal{L}_t$ will get arbitrarily close to $\log p(y_t \vert \vy_{1:t-1})$ as $N \to \infty$.
\end{proof}

\section{Proof of Corollary~\ref{corr:filtering_elbo}}\label{proof:corollary}
\begin{proof}
	The implicit \textit{smoothing} distribution that arises from performing SMC~\cite{Naesseth2018} is defined as
	\begin{equation}\label{eq:smc_smooth_implicit_dist}
		\tilde{q}(\vx_{1:t}) = p(\vx_{1:t}, \vy_{1:t})\mathbb{E}\left[\frac{1}{\hat{p}(\vy_{1:t})}\right].
	\end{equation}
	We can obtain the implicit filtering distribution by marginalizing out $d\vx_{1:t-1}$ from \eqref{eq:smc_smooth_implicit_dist}
	\begin{equation}\label{eq:smc_implici_dist}
		\begin{aligned}
			 & \tilde{q}(x_t \vert \vy_{1:t})                                                                                                          \\
			 & = \int  p(\vx_{1:t}, \vy_{1:t})\mathbb{E}\left[\frac{1}{\hat{p}(\vy_{1:t})}\right] d\vx_{1:t-1},                                       \\
			 & = p(x_{t}, \vy_{1:t})\mathbb{E}\left[\frac{1}{\hat{p}(\vy_{1:t})}\right],                                                              \\
			 & = p(x_t, y_t \vert \vy_{1:t-1}) \mathbb{E}\left[\hat{p}(y_ t \vert \vy_{1:t-1})^{-1}\frac{p(\vy_{1:t-1})}{\hat{p}(\vy_{1:t-1})} \right].
		\end{aligned}
	\end{equation}
	In~\cite{Naesseth2018,anh2018autoencoding}, it was shown that
	\begin{equation}\label{eq:elbo_lb}
		\begin{split}
			\log  p(\vy_{1:t})
			& \geq \mathbb{E}_{q(\vx_{1:t})}[\log p(\vx_{1:t}, \vy_{1:t}) - \log \tilde{q}(\vx_{1:t})]  \\
			& \geq   \mathbb{E}[\log \hat{p}(\vy_{1:t})].
		\end{split}
	\end{equation}
	Rearranging terms in~\eqref{eq:elbo_lb}, we get
	\begin{equation}\label{eq:double_lb}
		\log p(y_t \vert \vy_{1:t-1}) \geq \hat{\mathcal{L}}_t\geq \mathcal{L}_t.
	\end{equation}
	where
	\begin{equation}
		\begin{split}
			\hat{\mathcal{L}}_t & = \mathbb{E}_{q(\vx_{1:t})}[\log p(x_t, y_t \vert \vy_{1:t-1}, \vx_{1:t-1})  - \log \tilde{q}(x_t \vert \vx_{1:t-1})]  \\
			& + \KL[\tilde{q}(\vx_{1:t-1}) \Vert p(\vx_{1:t-1}, \vy_{1:t-1})] - \log p(\vy_{1:t-1}).
		\end{split}
	\end{equation}
	By Theorem \ref{thm:filtering_elbo}, we know that $\lim\limits_{N \to \infty} \mathcal{L}_t =\log p(y_t \vert \vy_{1:t-1})$, and thus
	\begin{equation}\label{eq:equal_elbo}
		\lim\limits_{N \to \infty} \hat{\mathcal{L}}_t = \log p(y_t \vert \vy_{1:t-1}).
	\end{equation}
	Leveraging Theorem 1 from~\cite{Naesseth2018} we have
	\begin{equation}
		\lim_{N \to \infty } \KL[\tilde{q}(\vx_{1:t-1}) \Vert p(\vx_{1:t-1}, \vy_{1:t-1})] = \log p(\vy_{1:t-1})
	\end{equation}
	which implies that
	\begin{equation}
		\lim_{N \to \infty } \tilde{q}(\vx_{1:t-1} ) = p(\vx_{1:t-1} ) \, \, \textrm{a.e.}
	\end{equation}
	thus plugging this into \eqref{eq:equal_elbo}
	\begin{equation}
		\begin{aligned}
			 & \log p(y_t  \vert \vy_{1:t-1})                                                                                                                                               \\
			 & = \int -\tilde{q}(\vx_{1:t}) \log \frac{p(x_t, y_t \vert \vy_{1:t-1}, \vx_{1:t-1})}{\tilde{q}(x_t \vert \vx_{1:t-1})} \, d\vx_{1:t}                                           \\
			 & =  \int -\tilde{q}(\vx_{1:t}) \log \frac{p(\vx_{1:t} \vert \vy_{1:t}) p(y_t \vert \vy_{1:t-1})}{\tilde{q}(x_t \vert \vx_{1:t-1}) p(\vx_{1:t-1}\vert \vy_{1:t-1})} \, d\vx_{1:t} \\
			 & = \log p(y_t \vert \vy_{1:t-1})                                                                                                                                              \\
			 & \qquad + \int -\tilde{q}(\vx_{1:t}) \log \frac{p(x_t \vert  \vx_{1:t-1}, \vy_{1:t})}{\tilde{q}(x_t \vert \vx_{1:t-1})} \, d\vx_{1:t}
		\end{aligned}
	\end{equation}
	which is true iff $\tilde{q}(x_t \vert \vx_{1:t-1}) = p(x_t \vert \vx_{1:t-1}, \vy_{1:t})$ almost everywhere.
	Thus by Lebesgue's dominated convergence theorem~\cite{van2000asymptotic}
	\begin{equation}
		\begin{split}
			& \lim_{N \to \infty } \int \tilde{q}(x_t \vert \vx_{1:t-1}) d\vx_{1:t-1} \\
			& = \int \lim_{N \to \infty }  \tilde{q}(x_t \vert \vx_{1:t-1}) d\vx_{1:t-1} \\
			& = p(x_t \vert \vy_{1:t}).
		\end{split}
	\end{equation}
\end{proof}

\section{Synthetic NASCAR\textsuperscript{\textregistered} Dynamics}\label{sec:supp:nascar}
An rSLDS~\cite{linderman2017bayesian} with 4 discrete states was used to generate the synthetic NASCAR\textsuperscript{\textregistered} track. The linear dynamics for each hidden  state were
\begin{align}
	A_1=
	\begin{bmatrix}
		\cos(\theta_1) & -\sin(\theta_1) \\
		\sin(\theta_1) & \cos(\theta_1)
	\end{bmatrix}
	,\,
	A_2=
	\begin{bmatrix}
		\cos(\theta_2) & -\sin(\theta_2) \\
		\sin(\theta_2) & \cos(\theta_2)
	\end{bmatrix}
	,
\end{align}
and $A_3 = A_4 = I$. The affine terms were $B_1 = -(A_1 - I)c_1$, ($c_1 = [ 2, 0]^\top$), $B_2 = -(A_2 - I)c_2$,  ($c_2 = [ -2, 0]^\top$), $B_3 = [0.1, 0]^\top$ and $B_4 = [-0.35, 0]^\top$. The hyperplanes, $R$, and biases, $r$, were defined as
\[R=
	\begin{bmatrix}
		100  & 0   \\
		-100 & 0   \\
		0    & 100
	\end{bmatrix}
	, \quad r=
	\begin{bmatrix}
		-200 \\
		-200 \\
		0
	\end{bmatrix} .
\]
A state noise of $Q = 0.001 I$ was used.

\section{Prediction using SVMC-GP}\label{sec:supp:prediction}
Let $\tilde{w}_t^i = \frac{w_t^i}{\sum_{\ell} w^\ell_t}$ be the self-normalized importance weights.
At time $t$, given a test point $x_*$ we can approximately sample from the predictive distribution
\begin{equation}\label{eq:gp_prediction}
	\begin{split}
		& p(f_* \vert x_*, \vy_{1:t}) \\
		& = \int p(f_* \vert x_*, \vz_t) p(\vz_t \vert \vy_{1:t}) d\vz_t \\
		& = \int p(f_* \vert x_*, \vz_t) p(\vz_t \vert \vx_{0:t}) p(\vx_{0:t} \vert \vy_{1:t}) d\vz_t d\vx_{0:t} \\
		& = \int p(f_* \vert x_*, \vx_{0:t}) p(\vx_{0:t} \vert \vy_{1:t}) d\vx_{0:t} \\
		& \approx \sum_{i=1}^N \tilde{w}_t^i \, p(f_* \vert x_*, \vx_{0:t}^i) \\
		&= \sum_{i=1}^N \tilde{w}_t^i \, \mathcal{N}(v^i_*, \Sigma^i_*)
	\end{split}
\end{equation}
where
\begin{align}
	v^i_*      & = m(x_*) + A_*\mu^i_t,        \\
	\Sigma^i_* & = C_* + A_*\Gamma^i_tA_*^\top
\end{align}
where $A_* = K_{*z}K_{zz}^{-1}$ and $C_* = K_{**} - K_{*z} K_{zz}^{-1} K_{z*} + Q$.
The approximate predictive distribution is a mixture of SGPs, allowing for a much more richer approximation to the predictive distribution. 
Equipped with~\eqref{eq:gp_prediction}, we approximate the mean of the predictive distribution, $\mu_{f_*}$, as
\begin{equation}\label{eq:predicted_mean}
	\begin{split}
		\mu_{f_*} & = \int f_* p(f_*\vert x_*, \vy_{1:t}) df_* \\
		& \approx \int f_* \sum_{i=1}^N \tilde{w}_t^i \,  p(f_* \vert x_*, \vx_{0:t}^i) df_* \\
		& = \sum_{i=1}^N \tilde{w}_t^i  \int f_* p(f_* \vert x_*, \vx_{0:t}^i) df_* \\
		& = \sum_{i=1}^N \tilde{w}_t^i \,  \mathbb{E}_i[f_*]
		= \sum_{i=1}^N \tilde{w}_t^i  v_*^i = \hat{\mu}_{f_*}
	\end{split}
\end{equation}
where $\mathbb{E}_i [\cdot]= \mathbb{E}_{p(f_* \vert x_*, \vx_{0:t}^i)}[\cdot] $.

Similarly, we can also approximate the covariance of of the predictive distribution, $\Sigma_{f_*}$
\begin{equation}
	\begin{aligned}
		 \Sigma_{f_*}
		 & = \int (f_* - \mu_{f_*}) (f_* - \mu_{f_*})^\top p(f_* \vert x_*, \vy_{1:t}) df_*                                                            \\
		 & \approx \sum_{i=1}^N \tilde{w}_t^i \int (f_* - \mu_{f_*}) (f_* - \mu_{f_*})^\top p(f_* \vert x_*, \vx_{0:t}^i) df_*                         \\
		 & = \sum_{i=1}^N \tilde{w}_t^i \, \mathbb{E}_i[(f_* - \mu_{f_*}) (f_* - \mu_{f_*})^\top] \\
		 & = \sum_{i=1}^N \tilde{w}_t^i (\mathbb{E}_i[f_* f_*^\top ] - \mathbb{E}_i[f_*] \mu_{f_*}^\top  - \mu_{f_*} \mathbb{E}_i[f_*]^\top + \mu_{f_*} \mu_{f_*}^\top )                 \\
		 & = \sum_{i=1}^N \tilde{w}_t^i (\Sigma_*^i + v_*^i v_*^{i\top} - v_*^i \mu_{f_*}^{\top} - \mu_{f_*} v_*^{i\top} + \mu_{f_*} \mu_{f_*}^\top)                          \\
		 & \approx \sum_{i=1}^N \tilde{w}_t^i (\Sigma_*^i + v_*^i v_*^{i\top} - v_*^i \hat{\mu}_{f_*}^{\top} - \hat{\mu}_{f_*} v_*^{i\top} + \hat{\mu}_{f_*} \hat{\mu}_{f_*}^\top).
	\end{aligned}
\end{equation}

\section{Winner-Take-All Spiking Neural Network}
\begin{figure}[bth]
	\centering
	\includegraphics[width=\linewidth]{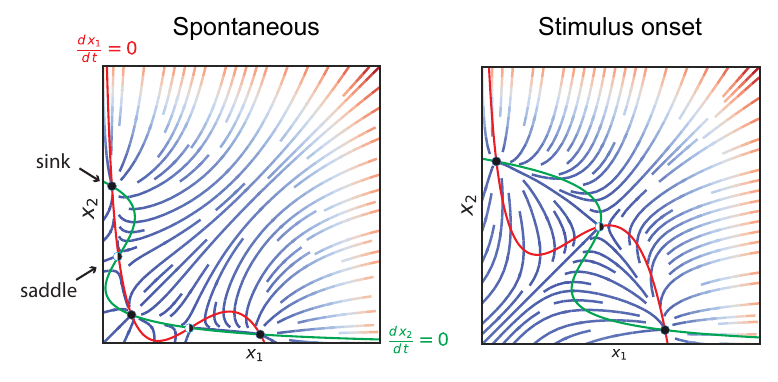}
	\caption[]{Mean field reduction of the Winner-Take-All spiking neural network.}
	\label{fig:wang_mean_field}
\end{figure}
In Figure~\ref{fig:wang_mean_field} the mean-field reduction of the spiking network model is shown~\cite{Wong2006}.

%
%

\ifCLASSOPTIONcaptionsoff
	\newpage
\fi



\bibliographystyle{IEEEtran}
\bibliography{ref.bib,catniplab.bib}
%
%
%

%
\begin{IEEEbiographynophoto}{Yuan Zhao}
	received his Ph.D. from Stony Brook University in 2016. He is a postdoc in the Department of Neurobiology and Behavior at Stony Brook University. His research interests lie in machine learning and computational neuroscience.
\end{IEEEbiographynophoto}
\vspace{-2ex}
\begin{IEEEbiographynophoto}{Josue Nassar}
	received the B.S. and M.S. degree in electrical engineering from Stony Brook University in 2016 and 2018, respectively.
	He is currently a Ph.D. candidate in the department of electrical and computer engineering at Stony Brook University.
	His research interest lie at the intersection of computational neuroscience, control, signal processing and machine learning.
\end{IEEEbiographynophoto}
\vspace{-2ex}
\begin{IEEEbiographynophoto}{Ian Jordan}
	received the BS degree in electrical engineering, specializing in control systems, from the New Jersey Institute of Technology in 2017.
	He is currently a PhD candidate at the Department of Applied Mathematics and Statistics at Stony Brook University.
	His research interests lie primarily in the field of applied dynamical systems theory, and the theory behind the underlying dynamics of recurrent neural networks.
\end{IEEEbiographynophoto}
\vspace{-2ex}
\begin{IEEEbiographynophoto}{M\'{o}nica~Bugallo}
M\'{o}nica F. Bugallo is a Professor of Electrical and Computer Engineering and Associate Dean for Diversity and Outreach of the College of Engineering and Applied Sciences at Stony Brook University. She received her B.S., M.S, and Ph. D. degrees in Computer Science and Engineering from University of A Coruña, Spain. Her research interests are in the field of statistical signal processing, with emphasis on the theory of Monte Carlo methods and its application to different disciplines including biomedicine, ecology, sensor networks, and finance. In addition, she has focused on STEM education and has initiated several successful programs with the purpose of engaging students at all academic stages in the excitement of engineering and research, with focus on underrepresented groups. She is a senior member of the IEEE and the Vice Chair of the IEEE Signal Processing Theory and Methods Technical Committee and has served on several technical committees of IEEE conferences and workshops.
\end{IEEEbiographynophoto}
\vspace{-2ex}
\begin{IEEEbiographynophoto}{Il Memming Park}
is an Associate Professor in Neurobiology and Behavior at Stony Brook University.
He is a computational neuroscientist trained in statistical modeling, information theory, and machine learning.
He received his B.S. in computer science from KAIST, M.S. in electrical engineering and Ph.D. in biomedical engineering from the University of Florida (2010), and trained at University of Texas at Austin as a postdoctoral fellow (2010-2014).
\end{IEEEbiographynophoto}




\end{document}